\documentclass[10pt,twocolumn,letterpaper]{article}

\usepackage[pagenumbers]{cvpr} %

\usepackage{graphicx}
\usepackage{amsmath}
\usepackage{amssymb}
\usepackage{booktabs}
\usepackage{cite}

\usepackage[pagebackref,breaklinks,colorlinks]{hyperref}

\usepackage[capitalize]{cleveref}
\crefname{section}{Sec.}{Secs.}
\Crefname{section}{Section}{Sections}
\Crefname{table}{Table}{Tables}
\crefname{table}{Tab.}{Tabs.}

\newcommand{\bK}{\mathbf{K}}

\newcommand{\bx}{\mathbf{x}}

\newcommand{\bI}{\mathbf{I}}

\newcommand{\bp}{\mathbf{p}}

\newcommand{\bD}{\mathbf{D}}

\newcommand{\bff}{\mathbf{f}}
\newcommand{\bF}{\mathbf{F}}

\newcommand{\bc}{\mathbf{c}}

\newcommand{\bd}{\mathbf{d}}

\newcommand{\bxi}{\boldsymbol{\xi}}

\newcommand{\nR}{\mathbb{R}}

\newcommand{\nS}{\mathbb{S}}

\newcommand{\cL}{\mathcal{L}}

\makeatletter
\DeclareRobustCommand\onedot{\futurelet\@let@token\@onedot}
\def\@onedot{\ifx\@let@token.\else.\null\fi\xspace}
\def\eg{e.g\onedot} 
\def\ie{i.e\onedot}

\def\etal{et~al\onedot} 
\def\Fig{Fig\onedot}   
\makeatother

\newcommand{\round}[1]{\ensuremath{\lfloor#1\rceil}}

\newcommand{\figref}[1]{\Fig~\ref{#1}}
\newcommand{\secref}[1]{Section~\ref{#1}}

\renewcommand{\eqref}[1]{Eq.~\ref{#1}}

\newcommand{\boldparagraph}[1]{\vspace{0.2cm}\noindent{\bf #1:} }

\newif\ifcomment
\commenttrue
\ifcomment
	\newcommand{\ag}[1]{ \noindent {\color{red} {\bf Andreas:} {#1}} }
	\newcommand{\yl}[1]{ \noindent {\color{cyan} {\bf Yiyi:} {#1}} }

\else
	\newcommand{\ag}[1]{}
	\newcommand{\yl}[1]{}
\fi

\usepackage{multirow}
\usepackage{makecell}
\usepackage{booktabs}

\def\confName{CVPR}
\def\confYear{2023}

\begin{document}

\title{SteerNeRF: Accelerating NeRF Rendering via Smooth Viewpoint Trajectory}

\makeatletter
\def\blfootnote{\gdef\@thefnmark{}\@footnotetext}
\makeatother

\author{Sicheng Li
\qquad  Hao Li 
\qquad Yue Wang 
\qquad Yiyi Liao$^{*}$
\qquad Lu Yu$^{**}$
\vspace{1em}
\\
Zhejiang University
}
\maketitle

\begin{abstract}
   Neural Radiance Fields (NeRF) have demonstrated superior novel view synthesis performance but are slow at rendering. To speed up the volume rendering process, many acceleration methods have been proposed at the cost of large memory consumption. To push the frontier of the efficiency-memory trade-off, we explore a new perspective to accelerate NeRF rendering, leveraging a key fact that the viewpoint change is usually smooth and continuous in interactive viewpoint control. This allows us to leverage the information of preceding viewpoints to reduce the number of rendered pixels as well as the number of sampled points along the ray of the remaining pixels. In our pipeline, a low-resolution feature map is rendered first by volume rendering, then a lightweight 2D neural renderer is applied to generate the output image at target resolution leveraging the features of preceding and current frames. We show that the proposed method can achieve competitive rendering quality while reducing the rendering time with little memory overhead, enabling 30FPS at 1080P image resolution with a low memory footprint. 
\end{abstract}
\blfootnote{$^*$ Corresponding author. $^{**}$ Co-corresponding author.}
\section{Introduction}

Novel View Synthesis (NVS) is a long-standing problem in computer vision and computer graphics with many applications in navigation~\cite{tancik2022block}, telepresence~\cite{zhang2022virtualcube}, and free-viewpoint video~\cite{xian2021space}. Given a set of posed images, the goal is to render the scene from unseen viewpoints to enable viewpoint control interactively.

With its ability to render high-fidelity images at novel viewpoints, Neural Radiance Fields (NeRF) have recently emerged as a popular representation for NVS. NeRF represents a scene as a continuous function, parameterized by a multilayer perceptron (MLP), that maps a continuous 3D position and a viewing direction to a density and view-dependent radiance~\cite{Mildenhall2020ECCV}. A 2D image is then obtained via volume rendering, \ie, accumulating colors along each ray.

However, the rendering process of NeRF is relatively slow as the MLP needs to be queried at millions of samples to render a single image, preventing NeRF from interactive view synthesis. Many recent works have focused on improving the rendering speed of NeRF, yet there is a trade-off between rendering speed and memory cost. State-of-the-art acceleration approaches typically achieve fast rendering at the expense of large memory consumption~\cite{HedmanSMBD21SNeRG}~\cite{YuLT0NK21plenoctree}, \eg, by pre-caching the intermediate output of the MLP, leading to hundreds of megabytes to represent a single scene.
While there are some attempts to accelerate NeRF rendering with a low memory footprint~\cite{kurz-adanerf2022}~\cite{NeffSPKMCKS21donerf}, the performance has yet to reach cache-based methods.
In practice, it is desired to achieve faster rendering at a lower memory cost. 

To push the frontier of this trade-off, we propose to speed up NeRF rendering from a new perspective, leveraging the critical fact that the viewpoint trajectory is usually smooth and continuous in interactive control. Different from existing NeRF acceleration methods that reduce the rendering time of each viewpoint \textit{individually}, we accelerate the rendering by exploiting the information overlap between \textit{multiple} consecutive viewpoints.

\begin{figure}[t]
  \centering
   \includegraphics[width=1.0\linewidth]{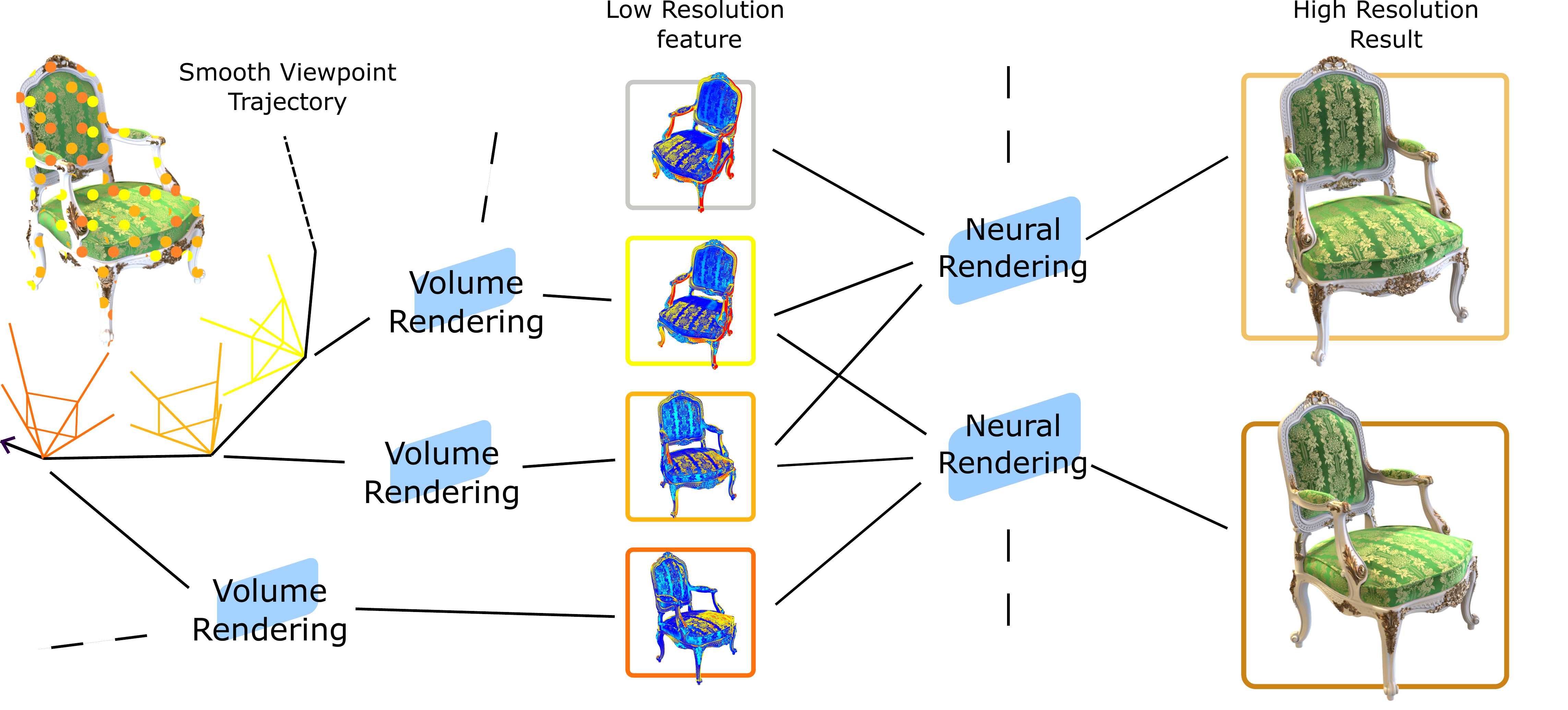}
   \caption{\textbf{Illustration}. We exploit smooth viewpoint trajectory to accelerate NeRF rendering, achieved by performing volume rendering at a low resolution and recovering the target image guided by multiple viewpoints. Our method enables fast rendering with a low memory footprint.} 
   \label{fig:teaser}
\end{figure}

\figref{fig:teaser} illustrates our SteerNeRF, a simple yet effective framework leveraging the \emph{S}moo\emph{T}h vi\emph{E}wpoint traj\emph{E}cto\emph{R}y to speed up NeRF rendering. 
Exploiting preceding viewpoints, we can accelerate volume rendering by reducing the number of sample points and maintain the image fidelity using efficient 2D neural rendering.

More specifically, 
our method consists of a rendering buffer, neural feature fields, and a lightweight 2D neural renderer. 
At a given viewpoint, we first render a low-resolution feature map via volume rendering. The sampling range along each ray is reduced by fetching a depth map from the rendering buffer and projecting it to the current view.
This effectively reduces the volume rendering computation as both the number of pixels and the number of samples for the remaining pixels are reduced.
Next, we combine preceding and current feature maps to recover the image at the target resolution using a 2D neural renderer, \ie, a 2D convolutional neural network. The neural feature fields and the 2D neural renderer are trained jointly in an end-to-end manner.
 
The combination of the low-resolution volume rendering and high-resolution neural rendering leads to lower rendering time compared to directly performing volume rendering at a high resolution. It maintains high fidelity and temporal consistency at a low memory cost.

Our method is inspired by existing super-sampling methods for real-time rendering~\cite{XiaoNCFLK20}. In contrast to these methods, we consider the more challenging NVS task without access to the perfect geometry. In this case, we demonstrate joint training of the neural feature fields and the neural renderer is crucial to achieve high image fidelity. 

We summarize our contributions as follows.
\begin{itemize}
    \item We provide a new perspective on NeRF rendering acceleration based on the assumption of smooth viewpoint trajectory. Our method is orthogonal to existing NeRF rendering acceleration methods and can be combined with existing work to achieve real-time rendering at a low memory footprint.
    \item To fully exploit information of preceding viewpoints, we propose a simple framework that combines low-resolution volume rendering and high-resolution 2D neural rendering. With end-to-end joint training, the proposed framework maintains high image fidelity. 
    \item Our experiments on synthetic and real-world datasets show that our method achieved a rendering speed of nearly 100 FPS at an image resolution of $800\times 800$ pixels and 30 FPS at $1920\times 1080$ pixels. It is faster than other low-memory NeRF acceleration methods and narrows the speed gap between low-memory and cache-based methods.
\end{itemize}

\section{Related Work}
\boldparagraph{Advances in NeRF}
Neural radiance fields~\cite{Mildenhall2020ECCV} have received significant attention with photorealistic novel view synthesis performance. Meanwhile, the vanilla NeRF has several limitations.

Many works have been conducted to address the limitations of NeRF, including unseen scene generalization~\cite{ChenXZZXY021}~\cite{yu2021pixelnerf}~\cite{liu2022neuray}~\cite{wang2021ibrnet}, dynamic scene representation~\cite{Pumarola2021CVPR}~\cite{LiSZGL0SLGNL22}~\cite{ParkSHBBGMS21}~\cite{PengDWZSZB21}~\cite{peng2021neural}~\cite{park2021nerfies}~\cite{li2021neural}, sparse view training~\cite{niemeyer2021regnerf}~\cite{dsnerf}, surface reconstruction~\cite{yariv2020multiview}~\cite{Neus}~\cite{VOLSDF}~\cite{oechsle2021unisurf}, and training acceleration~\cite{yu2021plenoxels}~\cite{DVGO}~\cite{Chen2022ECCV}. In addition to representing a single scene, NeRF is also demonstrated to have many applications in generative modeling~\cite{GRAFSchwarzLN020}~\cite{Niemeyer2020GIRAFFE}~\cite{piganChanMK0W21}~\cite{gu2021stylenerf}~\cite{eg3dChanLCNPMGGTKKW22}~\cite{DengYX022gram}, and robotics~\cite{niceslamZhu2022CVPR}~\cite{imapSucarLOD21}.

Another important question is accelerating the inference time of NeRF, which is critical for practical applications, \eg, interactive viewing control. Thus, many works focus on rendering acceleration.

\boldparagraph{NeRF Rendering Acceleration}
Existing NeRF Rendering acceleration methods could be categorized into two groups: one reduces the computation at each sample point, and one reduces the number of sampling points.

In the first category, one line of works reduce the computation by pre-caching the intermediate output of the MLP ~\cite{HedmanSMBD21SNeRG}~\cite{GarbinK0SV21fastnerf}~\cite{WadhwaniK22squeezenerf}~\cite{zhang2022digging}~\cite{WuLBWF22diver}~\cite{YuLT0NK21plenoctree}~\cite{wang2022fourier}~\cite{hu2022efficientnerf}~\cite{chen2022mobilenerf} or completely omit the network by representing the scene using a voxel grid~\cite{DVGO}~\cite{yu2021plenoxels}. During rendering, these methods retrieve the pre-stored information directly from the table instead of querying a deep network, thus accelerating the rendering.
Another line of works reduce the computation at each sample by replacing the large MLPs with smaller ones~\cite{ReiserPL021KiloNeRF}~\cite{esposito2022kiloneus}~\cite{wu2022scalable}, and maintains the image fidelity by using many small MLPs to represent a single scene. Despite achieving fast rendering, all these methods scarify memory over time, indicating the trade-off of rendering speed and memory cost.

The second category of methods speeds up rendering without increasing the memory cost. The core idea is to adaptively allocate different number of sampling points on each ray based on the content, thus reducing the number of sample points and accelerating rendering. Existing works in this area demonstrate that the number of sampling points can be effectively reduced while maintaining competitive rendering quality~\cite{NeffSPKMCKS21donerf}~\cite{kurz-adanerf2022}~\cite{piala2021terminerf}. 
Neural light fields based methods~\cite{wang2022r2l}~\cite{attal2022learning}, which are equivalent to reducing the number of sampling points to one, also fall into this category. However, simply reducing the number of sampling points has not yet achieved the same level of speed as tabulation-based methods.

Our method is compatible with these two types of methods as we reduce the time from a new perspective, i.e., by combining low-resolution volume rendering and high-resolution neural rendering aided by preceding frames. The low-resolution volume rendering of our method can benefit from existing acceleration methods. The additional memory cost of our method is small as our neural renderer is lightweight. More importantly, when combined with tabulation-based approaches for volume rendering, 
the resolution of voxel grid for pre-caching could be reduced sufficiently, since there is no need to render high-resolution content during the volume rendering stage. 

\boldparagraph{Superresolution for Rendered Content} 
A few research works exist that focus on leveraging superresolution techniques for rendering acceleration, especially content rendered from game engine. 
One of the representative work is NVIDIA DLSS~\cite{DLSS2}. DLSS is a technology tailored to increase the rendering frame rate while maintaining high fidelity. 
Even though its technical details have not been fully disclosed, in general, DLSS collects the raw low-resolution texture input, accurate motion vectors, and depth buffers from the rendering engine and outputs high-quality full-resolution content.
Xiao \etal~\cite{XiaoNCFLK20} proposed a method similar to DLSS. They achieve a new state of the art in super-resolving rendered videos with extreme aliasing by using a new temporal super resolution design, which takes texture, depth, motion vector from the game engine as input as well. Different from above works, our method addresses the more challenging NVS task, and thus cannot obtain high precision motion vector for superresolution.
Instead, we take a volume-rendered noisy depth map to warp the preceding frame to align with the current frame. 
Moreover, end-to-end joint training makes feature maps more expressive and enables 2D neural renderer to hallucinate high-fidelity images from current and warped feature. In the field of NeRF, NeRF-SR~\cite{WangWGZT022nerfsr} is proposed to generate higher resolution images with low-resolution supervisions leveraging the sub-pixel information across different views. Note that NeRF-SR is not applicable for real-time rendering.

\section{Method}
In this work, we propose fully exploiting the smoothly changing viewpoints to accelerate the rendering process of NeRF. In general, we achieve rendering acceleration by reducing the total number of 3D points that need to be queried in volume rendering for each frame. 

\begin{figure*}
  \centering
   \includegraphics[width=\linewidth]{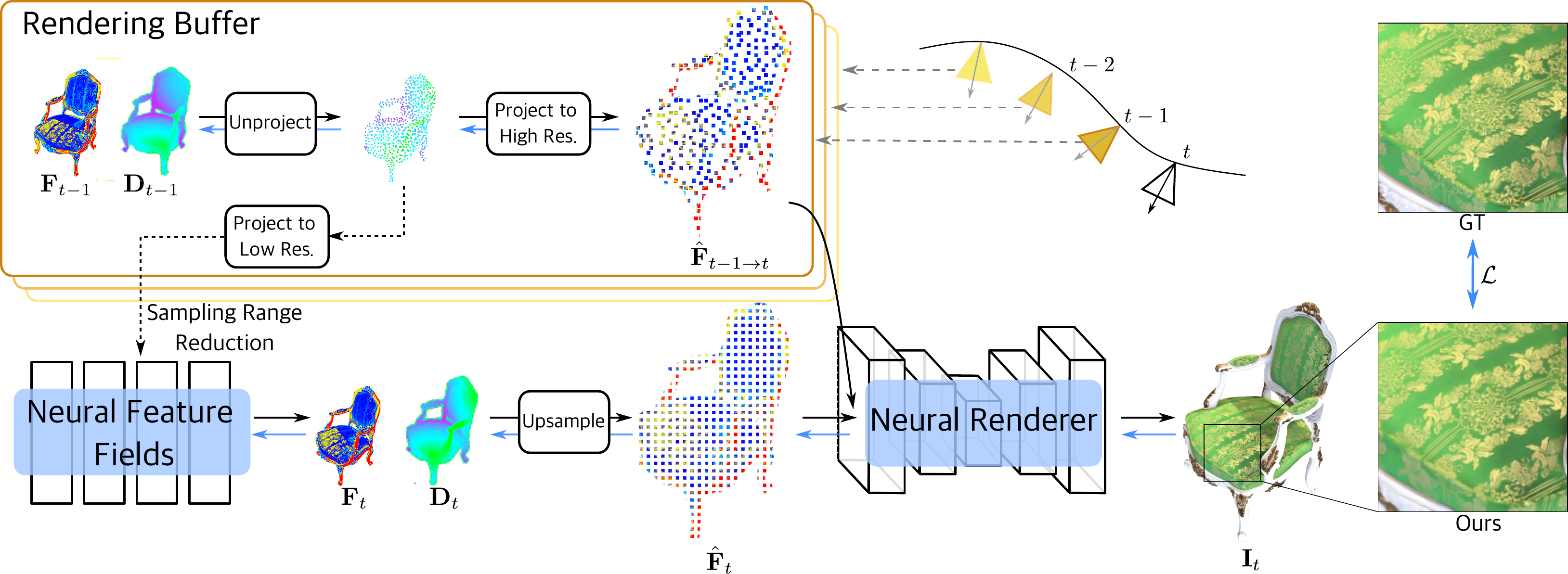}
   \caption{\textbf{SteerNeRF}. The rendering buffer saves low-resolution feature maps $\{\bF_{t-L}, \dots, \bF_{t-1}\}$ and depth maps $\{\bD_{t-L}, \dots, \bD_{t-1}\}$ of previous $L$ viewpoints. At the current viewpoint $t$, a low-resolution feature map $\bF_t$ and a depth map  $\bD_t$ are rendered accelerated by the rendering buffer. Next, the lightweight neural renderer takes as input the reprojected features maps at the high resolution $\{\hat{\bF}_{t-L \rightarrow t}, \dots, \hat{\bF}_{t-1 \rightarrow t}\}$ and the upsampled feature map $\hat{\bF_t}$ to generate the output image $\bI_t$. As illustrated by the blue arrows, during training, we apply the reconstruction loss $\cL$ to an image patch and jointly optimize the entire model in an end-to-end manner, including preceding frames in the rendering buffer. }
   \label{fig:pipeline}
\end{figure*}

\figref{fig:pipeline} gives an overview of our proposed pipeline consisting of a rendering buffer, neural feature fields, and a 2D neural renderer. 
Specifically, the rendering buffer saves low-resolution feature maps and depth maps of previous viewpoints. At the current viewpoint, a low-resolution feature map and depth map is rendered accelerated by the rendering buffer. Next, the lightweight neural renderer takes the preceding and the current feature maps as input to generate the output image at the target resolution.

In the following, we first introduce preliminaries of NeRF model in \secref{sec:nerf}. Next, we present the accelerated volume rendering in \secref{sec:acceleration}, the buffer-guided neural rendering in \secref{sec:unet}, and the training procedure in \secref{sec:training}. Finally, we describe implementation details in \secref{sec:detail}. 

\subsection{Background}
\label{sec:nerf}
NeRF represents a scene as a continuous function $f_\theta$ parameterized by learnable parameters $\theta$ that maps a 3D point $\bx\in \nR^3$ and a viewing direction $\bd \in \nS^2$ and to a volume density $\sigma$ and a color value $\bc$:
\begin{equation}
    f_{\theta}: (\bx\in\nR^3, \bd\in\nS^2) \mapsto (\sigma\in\nR^+,\bc\in\nR^3)
\end{equation}
Given a target viewpoint, the color $\bc_r$ and depth $d_r$ at a camera ray $r$ is obtained via volume rendering integral approximated by the numerical quadrature~\cite{max1995optical}:
\begin{gather}
    \bc_r =\sum_{i=1}^{N} T_r^i \alpha_r^i \bc_r^i \  \quad  d_r =\sum_{i=1}^{N} T_r^i \alpha_r^i  t_r^i \ \\
    \alpha_r^i = 1-\exp (-\sigma_r^i\delta_r^i) \quad  T_r^i   = \prod_{j=1}^{i-1}\left(1-\alpha_j\right)
\end{gather}
where $T_r^i$ and $\alpha_r^i$ denote transmittance and alpha value of a sample point $\bx_i$.

\boldparagraph{Rendering Time} The rendering time of NeRF is proportional to the amount of computation required to render an image. 
Let $H \times W$ denote the target image resolution,
$N$ the number of samples on each ray and $F$ the FLOPs of querying one sample's color and density.  We can roughly estimate the amount of computation as $H \times W \times N \times F $. 
Existing NeRF acceleration strategies mainly focus on how to decrease the FLOPs $F$ of each query by pre-caching the output of $f_\theta$ or directly use a voxel grid ~\cite{HedmanSMBD21SNeRG}~\cite{GarbinK0SV21fastnerf}~\cite{WadhwaniK22squeezenerf}~\cite{YuLT0NK21plenoctree}~\cite{yu2021plenoxels}, thus leading to large memory consumption. There are a few attempts to reduce the sampling points $N$ along the ray via early ray termination, empty space skipping~\cite{WuLBWF22diver} or adaptive sampling~\cite{kurz-adanerf2022}, yet using these techniques alone has not yet reached the performance of pre-cache based methods.

Our solution can elevate rendering speed from a new perspective, that is, reduce the number of pixels $H \times W$ and the number of samples $N$ for volume rendering via fully utilizing the smooth viewpoint trajectory. Besides, our solution can cooperate with the existing work to achieve high-speed rendering, leading to a higher rendering framerate while maintaining visual quality.

\subsection{Accelerating Volume Rendering}
\label{sec:acceleration}
We propose to learn a neural feature fields that renders a low-resolution feature map suited for the subsequent neural renderer. The acceleration of our framework comes from 1) rendering the feature map at a lower resolution and 2) reducing the sampling range guided by the rendering buffer. 

\boldparagraph{Low-Resolution Feature Rendering}
We render a feature map via volume rendering.
Our neural feature fields maps the input $\bx$ and the viewing direction $\bd$ to a density value and a feature vector $\bff$:
\begin{equation}
    f_{\theta}: (\bx\in\nR^3, \bd\in\nS^2) \mapsto (\sigma\in\nR^+,\bff \in\nR^K)
\end{equation}
where $K$ is the number of channels of our feature vector. 
Despite providing more information, rendering extra channels leads to a very litte overhead in time as we only expand the last layer of the MLP to predict more channels.

We can obtain a feature vector $\bff_r$ at each ray $r$ via volume rendering. 
\begin{equation}
     \bff_r =\sum_{i=1}^{N} T_r^i \alpha_r^i \bff_r^i 
\end{equation}
We render the feature vector at a subset of the rays, yielding a feature map $\bF \in \nR^{\frac{H}{4}\times\frac{W}{4}\times K} = \{\bff_{r}\}$.
The corresponding low-resolution depth value $\bD\in \nR^{\frac{H}{4}\times\frac{W}{4}} $ is also rendered. 
Both the feature map $\bF$ and $\bD$ are stored into our rendering buffer for subsequent frames.

\boldparagraph{Buffer-Guided Sampling Range Reduction}
The depth information of adjacent frames rendered previously provides a coarse scene geometry. Thus, warped depth from the preceding frame could be used as guidance to determine sampling positions and thus accelerate rendering. Specifically, given a low-resolution depth map $\bD_{t', t'<t}$ of the previous frame, it is first unprojected to a 3D point cloud and then projected to the current viewpoint. More formally,
the following unprojection and projection functions are applied to each pixel $(u,v)$ of $\bD_{t'}$ with depth $d_{t'}$:
\begin{align}
& \bp  = \bxi_{t'}^{-1} \bK_l^{-1} d_{t'}[u,v,1]^T \label{eq:unproject}\\
& d_{{t'}\rightarrow t}[u_{{t'}\rightarrow t},v_{{t'}\rightarrow t},1]^T = \bK_l \bxi_{t} \bp
\label{eq:warp_lowres}
\end{align}
where $\bp$ denotes a 3D point and $\bK_l$ is the intrinsic matrix of the low-resolution image. This yields the reprojected depth map $\bD_{{t'}\rightarrow t}$ where 
\begin{equation}
\bD_{{t'}\rightarrow t}(\round{u_{{t'}\rightarrow t}},\round{v_{{t'}\rightarrow t}})=d_{{t'}\rightarrow t}.
\label{eq:round}
\end{equation}
Note that here we simply round  $(u_{{t'}\rightarrow t},v_{{t'}\rightarrow t})$ and observe negligible impact on the performance. Given $\bD_{{t'}\rightarrow t}$, the sampling range at frame $t$ can be limited to the depth interval $[\bD_{t' \rightarrow t}-\epsilon,\bD_{{t'}\rightarrow t}+\epsilon]$ for rendering $\bF_t$ and $\bD_t$ at the camera viewpoint $\bxi_t$.
 This simple strategy further decreases the number of 3D sample points and accelerates the rendering of the low-resolution feature map.

\subsection{Buffer-Guided Neural Rendering}
\label{sec:unet}
Given the rendering buffer consisting of $L$ preceding feature maps $\{\bF_{t-L},\dots,\bF_{t-1}\}$ and depth maps  $\{\bD_{t-L},\dots,\bD_{t-1}\}$, we combine them with the feature map at the current viewpoint $t$ to recover the output image $\bI_t$ using a 2D neural renderer.
We first project frames in the rendering buffer to the current viewpoint. Next, we use a 2D neural renderer to recover the target image.

\boldparagraph{Preceding Frames Reprojection}
Inspired by natural video superresolution approaches, our method warps previous frames to align with the current frame to ease the task of the subsequent neural renderer. 
Instead of reprojecting the depth map to the low-resolution image as in \eqref{eq:warp_lowres}, we directly project the 3D point cloud to the target resolution to achieve higher precision, \ie, maintain sub-pixel precision in terms of the low-resolution image:
\begin{equation}
d_{{t'}\rightarrow t}[\hat{u}_{{t'}\rightarrow t},\hat{v}_{{t'}\rightarrow t},1]^T = \bK_h \bxi_{t} \bp
\label{eq:warp_highres}
\end{equation}
where $\bK_h$ denotes the intrinsic matrix of the high-resolution image. This allows us to obtain the reprojected high-resolution feature map $\hat{\bF}_{t'\rightarrow t}\in \nR^{H\times W \times K}$:
\begin{equation}
\hat{\bF}_{{t'}\rightarrow t}(\round{\hat{u}_{{t'}\rightarrow t}},\round{\hat{u}_{{t'}\rightarrow t}})=\bF_{t'}(u,v).
\label{eq:round}
\end{equation}

\boldparagraph{Neural Renderer}
We use a lightweight 2D convolutional network for fast inference. The reprojected high-resolution feature maps $\{\hat{\bF}_{{t'}\rightarrow t}\}$ are concatenated with the upsampled feature map $\hat{\bF}_t$ and mapped to the output target image:
\begin{equation}
    g_{\theta}: (\{\hat{\bF}_{{t'}\rightarrow t}\}, \hat{\bF}_t) \mapsto \bI_t \in \nR^{H\times W \times 3}
\end{equation}
In practice, we choose a simple modified U-Net as our neural renderer. Compared to traditional U-Net, we reduce the number of convolution layers for high-resolution features and increase the depth of convolution layers for low-resolution features. The simple adjustment allows us to greatly reduce the inference time and keep visual quality when the number of parameters is almost the same.
We use an off-the-shelf inference acceleration toolbox, NVIDIA TensorRT, to optimize neural renderer to reduce the inference time.

\subsection{Training}
\label{sec:training}
The training strategy is crucial to achieving high-quality novel view synthesis. In practice, we first pre-train our neural feature fields and then train the full model jointly in an end-to-end training fashion.

\boldparagraph{Pre-training}
We pre-train our neural feature fields on the target resolution $H\times W$. Here, we render a high-resolution feature map $\tilde{F}\in\nR^{H\times W\times K}$ and apply an $L2$ reconstruction loss on the first three channels supervised by the high-resolution ground truth image.
We leave other output channels without constraint of supervision as pretraining on the first three channels is sufficient to learn reasonable volume density.

\boldparagraph{End-to-end Joint Training}
With the pre-trained neural feature fields, we train our full model in an end-to-end manner using an $L2$ loss $\cL$ on the final output image $\bI\in\nR^{H\times W\times 3}$. Note that we do not apply reconstruction loss to the rendered feature map during end-to-end training and let the neural feature fields learn features suited for the 2D neural renderer. During training, the loss $\cL$ is applied to image patches.
As the training viewpoints are scattered in the space without a smooth trajectory, we generate a short sequence of preceding camera poses for each training image to train the buffer-based neural renderer. Note that this process does not introduce additional supervision as we only apply the loss $\cL$ to the training viewpoints despite taking preceding feature maps as input.

\subsection{Implementation Details}
\label{sec:detail}
\boldparagraph{Network Architecture}
Our method is compatible with different NeRF approaches for learning the neural feature fields. In this work, we implement our neural feature fields based on Instant-NGP~\cite{muller2022instant} using a third-party PyTorch implementation\footnote{\url{https://github.com/kwea123/ngp_pl}}. This allows more efficient feature map rendering than the vanilla NeRF. 
We follow the original architecture of Instant-NGP that uses multi-resolution hash tables where the table length at each resolution is fixed to $2^{19}$.
Following Instant-NGP, empty space skipping and early ray termination are applied when rendering the low-resolution feature map.
Regarding the 2D neural renderer, we adopt a shallow U-Net~\cite{ronneberger2015u} with the detailed architecture described in the supplementary.

\boldparagraph{Distillation}
When the number of training views is relatively small, the 2D neural renderer tends to overfit the training views, yielding degenerated performance on the test poses. In this case,
we leverage a pre-trained NeRF model to synthesize more viewpoints as our pseudo ground truth by randomly sampling viewpoints within the available viewing zone. Adding the randomly sampled pseudo ground truth  alleviates the overfitting problem.

\boldparagraph{Inference optimization}
Optimizing trained neural renderer for real-time inference and lower memory footprint is necessary. Thus, we leverage NVIDIA TensorRT to optimize 2D neural renderer. Prior to testing, we optimize 2D neural renderer into two versions in FP16 and INT8 precision separately.

\section{Experiments}
\begin{figure*}[t!]
     \setlength{\tabcolsep}{0.2pt}
     \def\mywidth{.5}
     \begin{tabular}{cc}
      \includegraphics[width=\mywidth\linewidth]{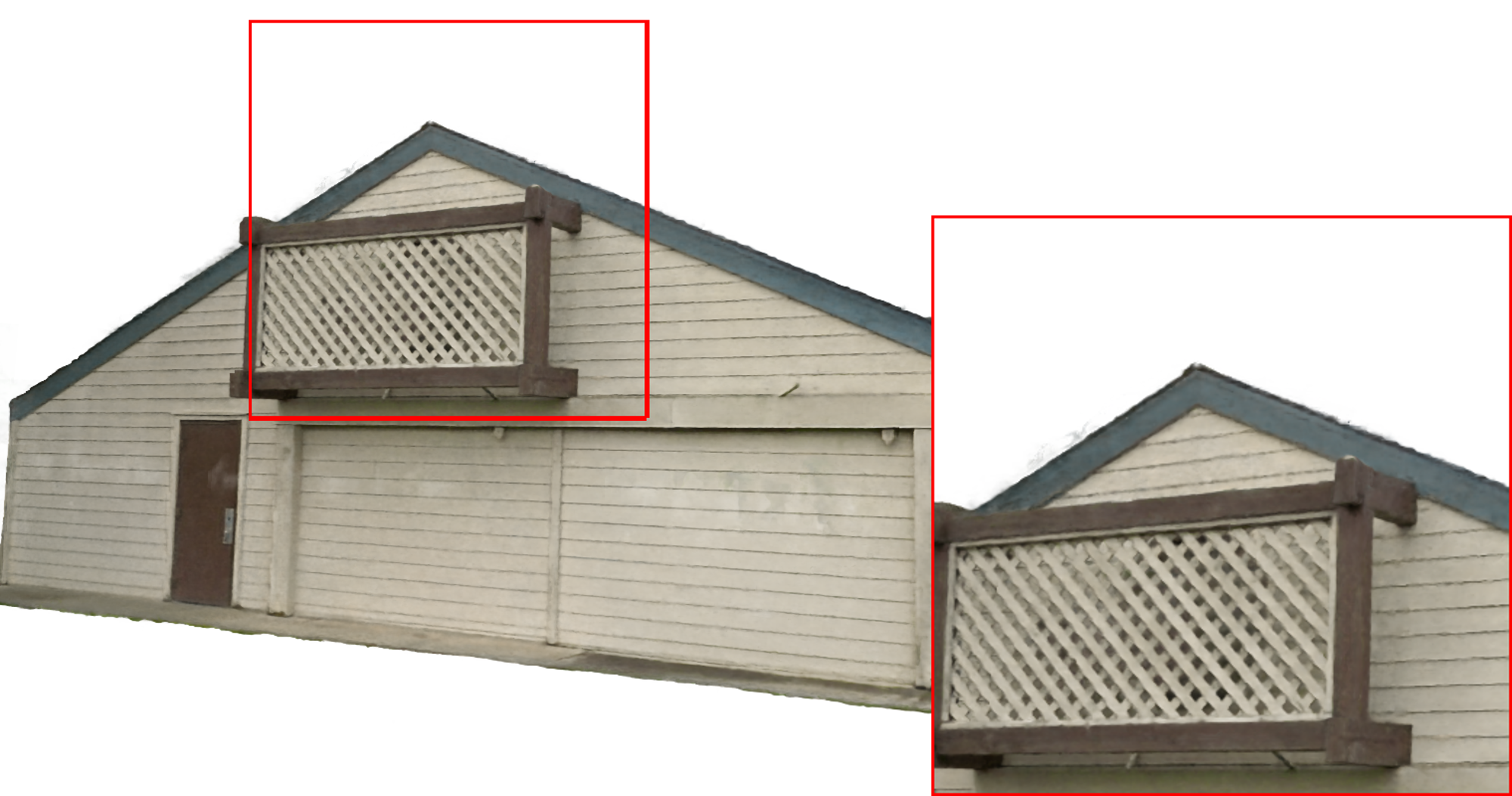} &
      \includegraphics[width=\mywidth\linewidth]{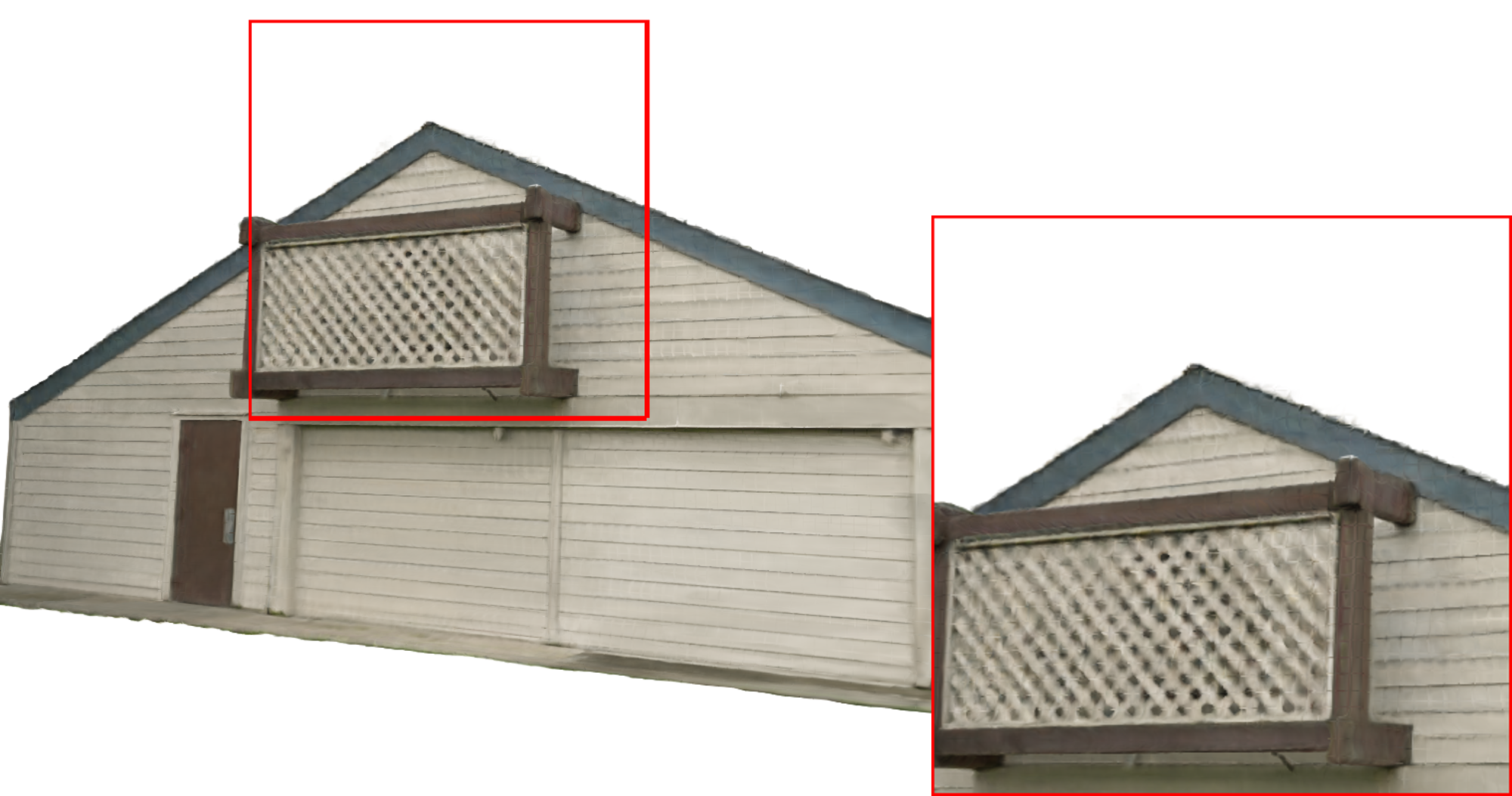} \\
      \begin{small}Instant-NGP~\cite{muller2022instant}\end{small} &
      \begin{small}DIVeR~\cite{WuLBWF22diver}\end{small} \\
      \includegraphics[width=\mywidth\linewidth]{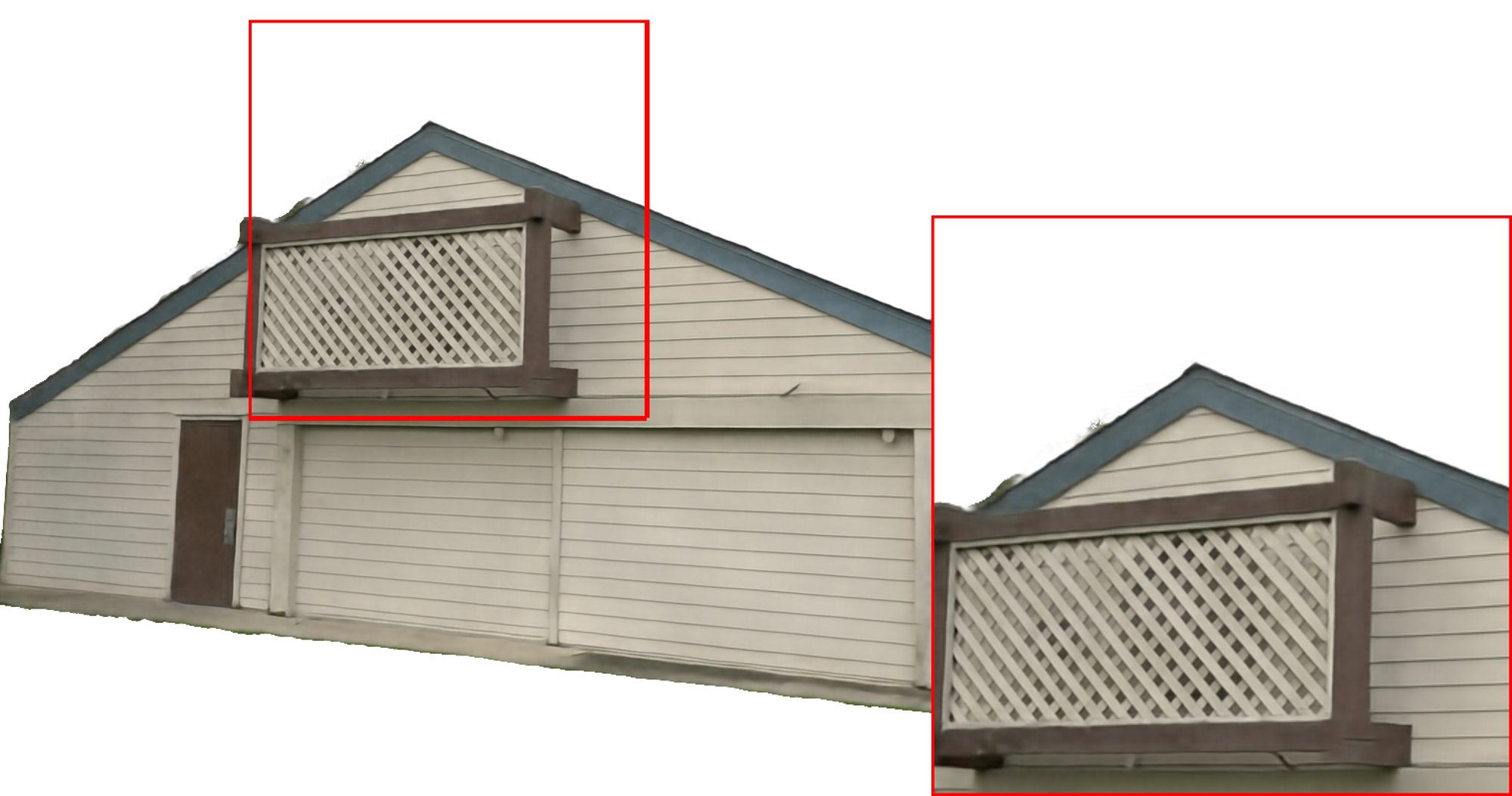} &
      \includegraphics[width=\mywidth\linewidth]{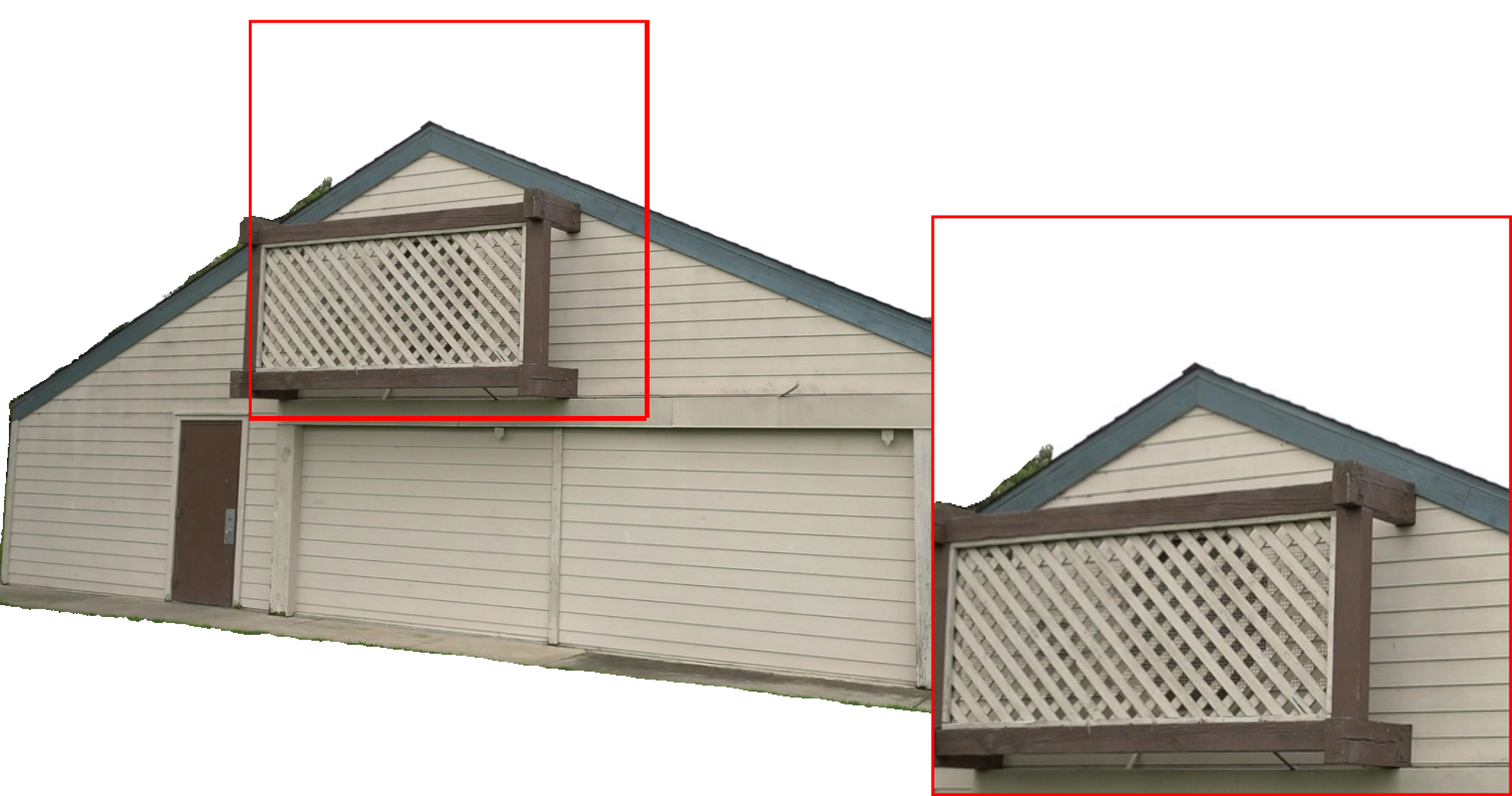} \\
      \begin{small}Ours\end{small} &
      \begin{small}GT\end{small} 
     \end{tabular}\vspace{-0.1cm}
     \caption{{\bf Qualitative Comparison on Tanks \& Temples.}  
     }
     \label{fig:tat}
     \vspace{-0.3cm}
    \end{figure*}
In this section, we evaluate the performance of our method. We first quantitatively compare our work with prior work. Next, we report the runtime breakdown of our method in two representative scenes. Finally, we validate our design decisions with extensive ablation studies. 
\begin{table*}[h!]
\centering

\begin{tabular}{@{}lccccccccc@{}}
\toprule
\multirow{2}{*}{Method} & \multicolumn{4}{c}{Tanks \& Temples} & \multicolumn{4}{c}{NeRF-Synthetic} & \multirow{2}{*}{Mem.(MB)$\downarrow$} \\
            & PSNR(dB)$\uparrow$  & SSIM$\uparrow$ & LPIPS$\downarrow$ & FPS$\uparrow$  & PSNR(dB)$\uparrow$ & SSIM$\uparrow$ & LPIPS$\downarrow$ & FPS$\uparrow$    &      \\ \midrule
NeRF~\cite{Mildenhall2020ECCV}        & 28.32 & 0.890 & 0.198 & 0.005 & 31.01 & 0.947 & 0.081 & 0.02   & \textbf{5}    \\
NSVF~\cite{LiuGLCT20nsvf}               & 28.40 & 0.900 & 0.153 & 0.06  & 31.74 & 0.953 & 0.047 & 0.23   & -    \\ \midrule
KiloNeRF~\cite{ReiserPL021KiloNeRF}    & 28.41 & 0.900 & \textbf{0.092} & 10.95 & 31.00 & 0.950 & \textbf{0.030} & 38.50  & 161  \\
PlenOctree~\cite{YuLT0NK21plenoctree}  & -     & -     & -     & -     & 31.71 & \textbf{0.958} & 0.053 & \textbf{167.7} & 1930 \\
DIVeR~\cite{WuLBWF22diver}              & 28.18 & 0.912 & \underline{0.116} & -     & \underline{32.12} & \textbf{0.958} & \underline{0.033} & 74.00  & 68   \\
Instant-NGP~\cite{muller2022instant} & \textbf{28.77} & 0.918 & 0.136 & 5.00  & \textbf{32.79} & \underline{0.957} & 0.055 & 60.00  & \underline{25.2} \\ \midrule
Ours (FP16)                            & \underline{28.65} & \textbf{0.924} & 0.121 & \underline{27.24} & 31.60 & 0.954 & 0.058 & 75.19  & 29.4 \\
Ours (INT8)                             & 28.44 & \underline{0.919} & 0.129 & \textbf{30.90} & 30.97 & 0.948 & 0.065 & \underline{86.97}  & 27.3 \\ \bottomrule
\end{tabular}
\caption{\textbf{Quantitative results on Tanks \& Temples and NeRF-Synthetic} show that our method could achieve high framerate rendering while keeping relatively low memory footprint. (\textbf{Best}, \underline{Second Best})}
\label{baseline}
\end{table*}

\boldparagraph{Datasets}
We take NeRF-Synthetic dataset, and a subset of the Tanks \& Temples dataset~\cite{knapitsch2017tanks} for performance evaluation. NeRF-Synthetic dataset contains  8 synthetic scenes rendered by Blender at a resolution of $800 \times 800$ and full of high frequency texture. Tanks \& Temples dataset is a real-world dataset at a resolution of $1920 \times 1080$. We follow the subset selection and crop the backgrounds of images as in 
NSVF~\cite{LiuGLCT20nsvf}.

\boldparagraph{Baselines}
The baseline methods could be categorized into two groups: one is classical but unable to achieve real-time rendering, and another is capable of real-time rendering. As for classical methods, we compare with the original NeRF and NSVF. As for real-time rendering methods, we compare with PlenOctree, DIVeR, and Instant-NGP. Note that we take DIVeR32 (RT) for a fair comparison, a real-time version of DIVeR with better speed-quality trade-off. 

\boldparagraph{Metrics}
We evaluate our method from efficiency, quality, and memory usage. The efficiency is measured by the number of frames per second. The quality is measured by PSNR, SSIM~\cite{Wang2004TIP} and LPIPS~\cite{ZhangIESW18lpips}. The memory usage is measured by megabytes.

\begin{figure*}[h]
    \setlength{\tabcolsep}{0pt}
    \def\mywidth{.25}
    \begin{tabular}{cccc}
    \includegraphics[width=\mywidth\linewidth]{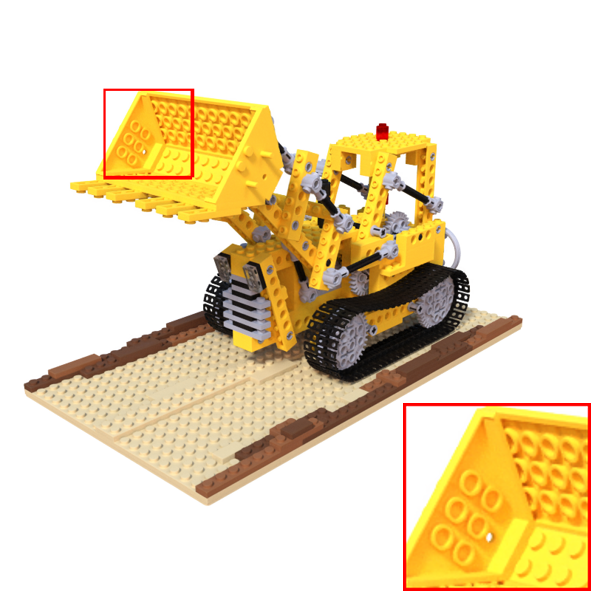} &
    \includegraphics[width=\mywidth\linewidth]{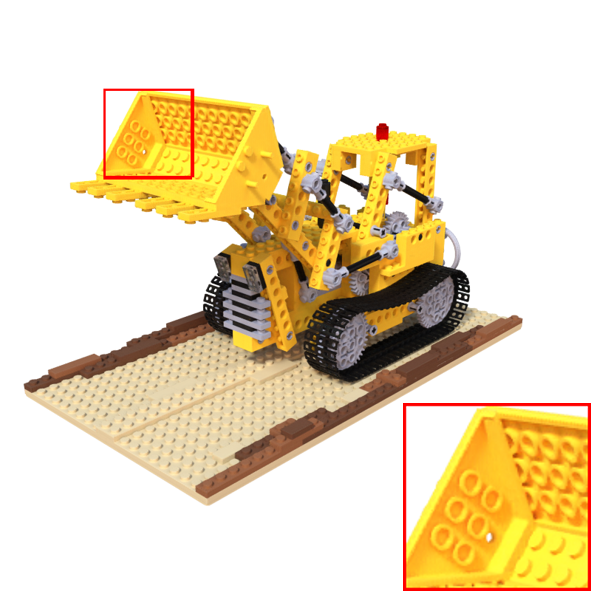} &
    \includegraphics[width=\mywidth\linewidth]{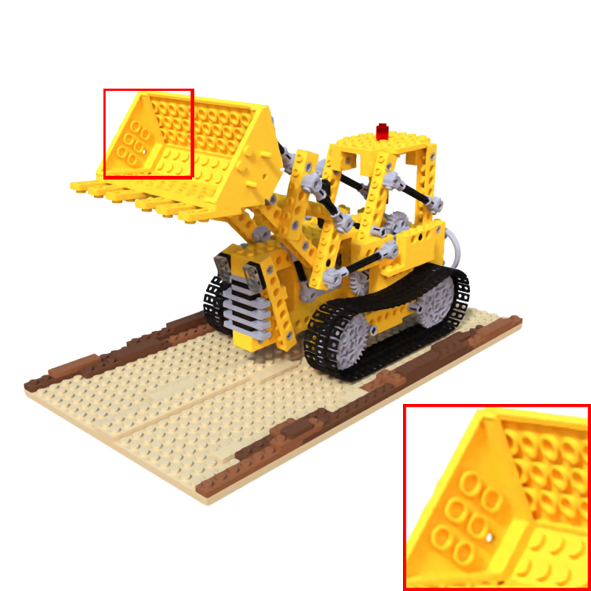} &
    \includegraphics[width=\mywidth\linewidth]{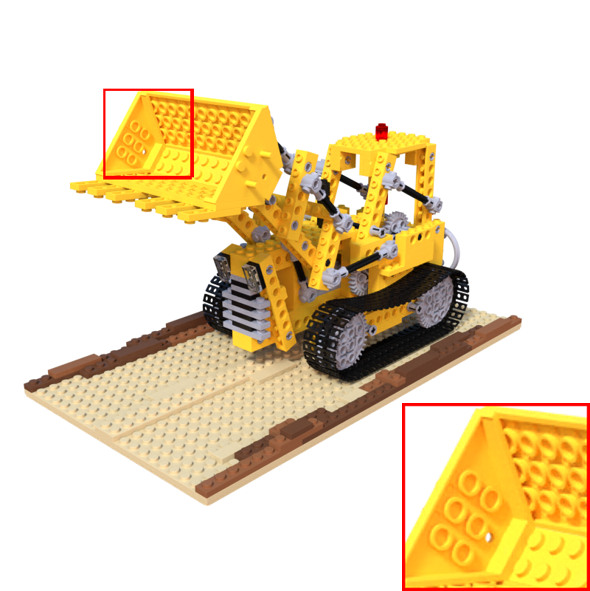} \\
    \includegraphics[width=\mywidth\linewidth]{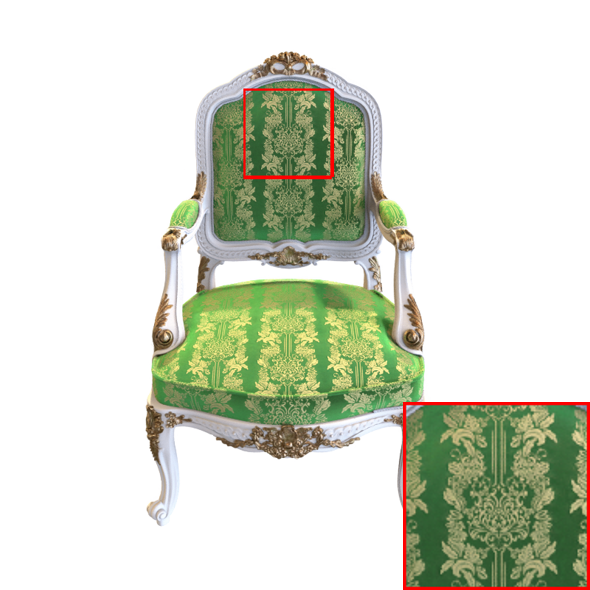} &
    \includegraphics[width=\mywidth\linewidth]{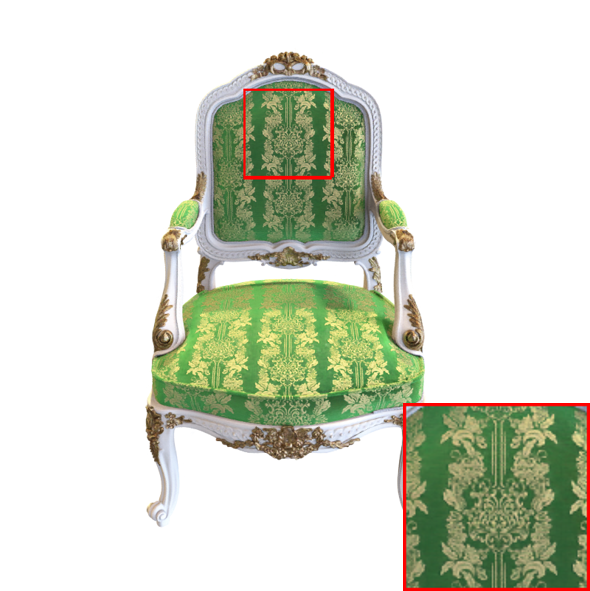} &
    \includegraphics[width=\mywidth\linewidth]{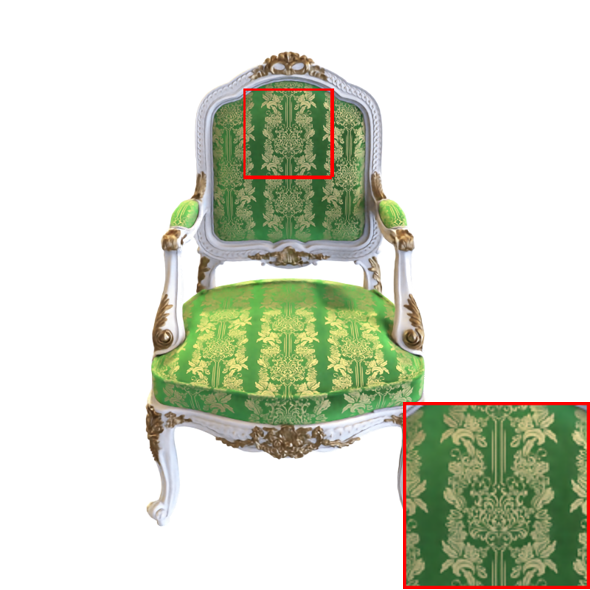} &
    \includegraphics[width=\mywidth\linewidth]{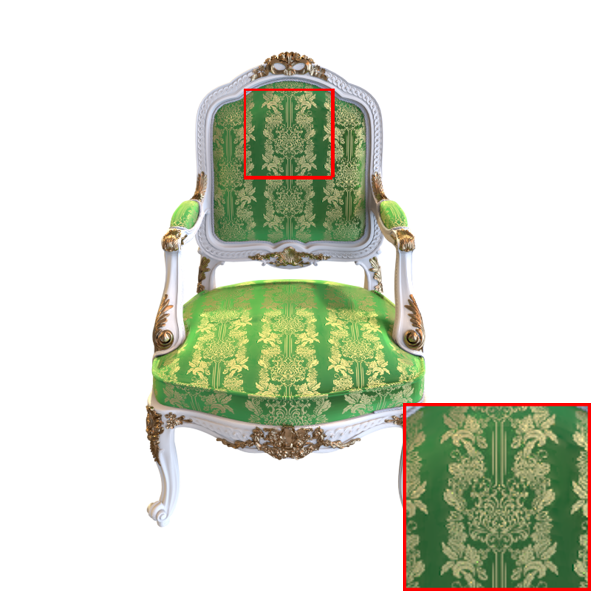} \\
    \includegraphics[width=\mywidth\linewidth]{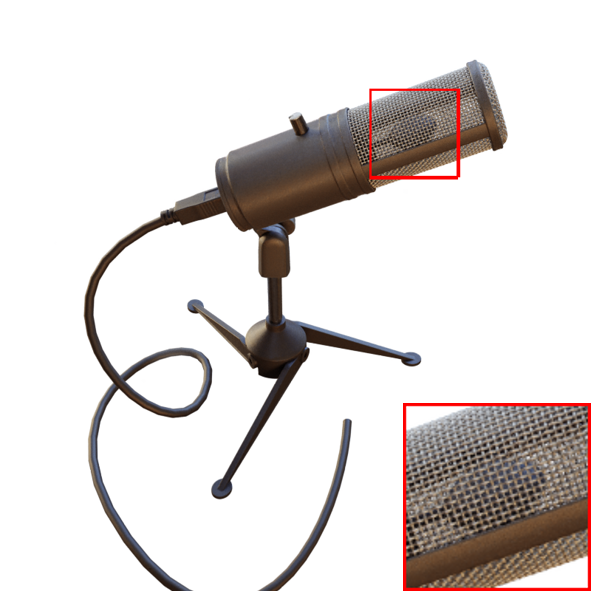} &
    \includegraphics[width=\mywidth\linewidth]{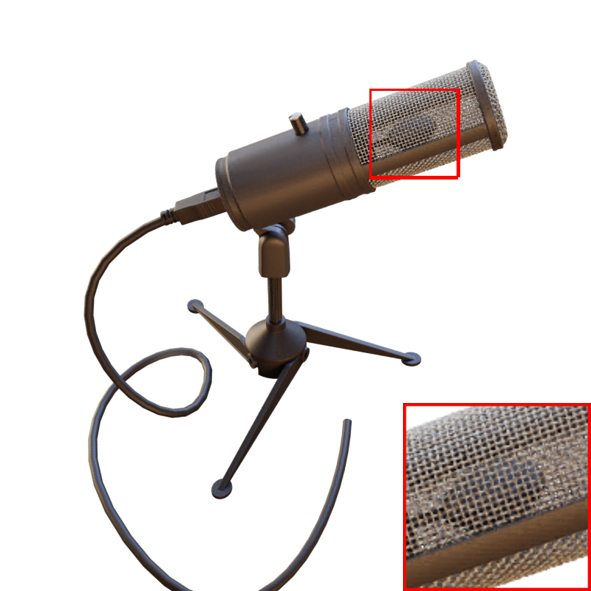} &
    \includegraphics[width=\mywidth\linewidth]{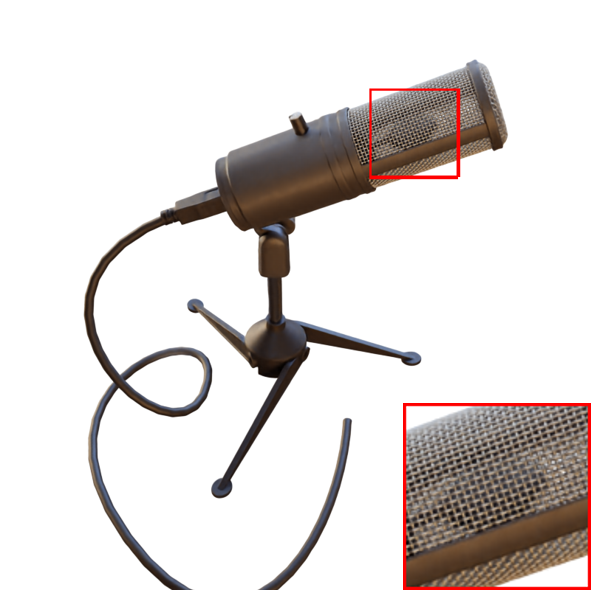} &
    \includegraphics[width=\mywidth\linewidth]{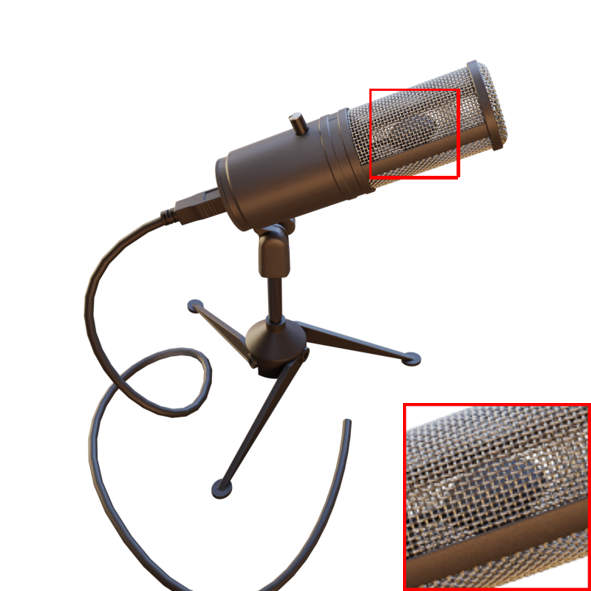} \\
    \begin{small}Instant-NGP~\cite{muller2022instant}\end{small} &
    \begin{small}DIVeR~\cite{WuLBWF22diver}\end{small} &
    \begin{small}Ours\end{small} &
    \begin{small}GT\end{small} 
    \end{tabular}\vspace{-0.1cm}
    \caption{{\bf Qualitative Comparison on NeRF-Synthetic.}  
    }
    \label{fig:synthetic}
    \vspace{-0.3cm}
\end{figure*}

\subsection{Comparisons to Baseline}

We first report the results on Tanks \& Temples dataset in Tab.~\ref{baseline} (left). All baseline and our methods show similar visual quality. However, our method shows the best FPS, almost three times higher than the best baseline method, KiloNeRF. In the meantime, the memory usage of our method is less than 20\%  of KiloNeRF. Instant-NGP, which has similar memory usage with ours, can only render views at 5 FPS. 
PlenOctree is not compared here as we fail to re-train the model to achieve reasonable results.
NeRF and NSVF, both employing a large MLP, perform much slower in rendering efficiency. With such a trade-off between memory usage and 
rendering efficiency, our method satisfies 3D interaction along smooth viewport trajectories at 1080P resolution with 30FPS. We show corresponding qualitative results on Tanks \& Temples in Fig.~\ref{fig:tat}.

On NeRF-Synthetic dataset in Tab.~\ref{baseline} (right), all methods still show similar visual quality scores. For FPS, PlenOctree becomes the best with one to two orders of magnitude higher memory usage, which is mainly caused by its network-free nature. Among the others, all the methods keep the same memory usage, and our method still shows a superior FPS. There are two differences from the previous dataset in FPS. First, KiloNeRF is slower than Instant-NGP. Second, the margin of our method against the others is smaller. We explain both differences by the bandwidth of GPU, which may not be fully maximized due to the lower image resolution i.e. fewer times of ray marching or neural network inference. In summary, we consider that the parallelism of our method is able to push the frontier of the efficiency-memory trade-off, especially for high resolution rendering. We present corresponding qualitative results on NeRF-Synthetic in Fig.~\ref{fig:synthetic}.

\subsection{Runtime Breakdown}
\begin{table}[h]
\small
\centering
\begin{tabular}{@{}lccccc@{}}
\toprule
\def\mywidth{0.01}

\multirow{2}{*}{Module}   & \multicolumn{5}{c}{Time(ms)}                              \\
                          & \multicolumn{2}{c}{Barn} &    & \multicolumn{2}{c}{Chair}    \\ \midrule
$\{\bD_{{t'}\rightarrow t}\}_{t'=t-1}$  & 0         & 0.11          &           & 0             & 0.11 \\ 
$f_\theta$                              & 20.28      & 17.15          &           &  4.81       & 3.76         \\ \midrule
$\{\hat{\bF}_{{t'}\rightarrow t}\}_{t'=\{t-2, t-1\}}$    
                                        & \multicolumn{2}{c}{3.21}  &           & \multicolumn{2}{c}{2.40}    \\ 
$g_\theta$                  & \multicolumn{2}{c}{12}              &           & \multicolumn{2}{c}{3.71}   \\ \midrule
Total                                   & 35.49       & 32.47       &           & 10.92         & 9.98         \\ \bottomrule
\end{tabular}   

\caption{\textbf{Runtime breakdown} for Chair and Barn scenes}
\label{bd}
\end{table}

We report the average runtime of our method in Tab.~\ref{bd} for Chair and Barn scenes, including the runtime of volume rendering, neural rendering and depth reprojection. For volume rendering,  $\{\bD_{{t'}\rightarrow t}\}_{t'=t-1}$ refers to the depth reprojection to enables buffer-guided sampling range reduction. When this fuction is turned on, it reduces the volume rendering time ($f_\theta$) to  85-90\% of the original. As for the neural rendering part, the reprojection $\{\hat{\bF}_{{t'}\rightarrow t}\}_{t'=\{t-2, t-1\}}$ takes longer as two frames are reprojected to a higher resolution. The reprojection can be further accelerated in the future work by using custom CUDA kernels.

\subsection{Ablation Study}
We conduct ablation studies on SteerNeRF using Chair scene from NeRF-Synthetic dataset.

\boldparagraph{Number of Preceding Frames}
We first evaluate the visual quality and render time by taking a varying number of preceding frames $L$ as input to neural renderer in Tab.~\ref{prev}. As more preceding frames are used, the reconstruction quality increases with a growing rendering time. The additional runtime mainly comes from the warping operation of preceding frames. Furthermore, we also observe that as the number of preceding frames increases, the quality gain brought by each additional frame decreases. Therefore, in practical applications, we can adjust this parameter more flexibly according to the scene content to achieve trade-off between quality and rendering efficiency.
\begin{table}[t]
\small
\centering

\begin{tabular}{@{}lcccc@{}}
\toprule
 $L$: \# Previous frames & 0      & 1      & 2      & 3   \\ \midrule
PSNR(dB)           & 32.94  & 33.05  & 33.14  & 33.10     \\ 
SSIM               & 0.959  & 0.964  & 0.968  &  0.967    \\ 
LPIPS              & 0.035  & 0.034  & 0.034  &  0.032    \\ 
Runtime(ms)        & 8.46   & 9.8    & 11.14  &  12.48    \\ \bottomrule
\end{tabular}
\vspace{-0.2cm}
\caption{\textbf{Comparison of Number of Preceding Frames} on Chair.}
\label{prev}
\end{table}

\boldparagraph{Number of Feature Channels}
We also verify the effect of the number of channels of feature images. Experiments show that as the number of channels increases, the gain of visual quality decreases gradually in Tab.~\ref{fc}. Therefore, we ended up choosing six channels in total for our implementation. Note that employing feature rendering almost causes no rendering time overhead.

\begin{figure}[t!]
 \setlength{\tabcolsep}{0.3pt}
 \def\mywidth{.25}
 \begin{tabular}{cccc}
  \includegraphics[width=\mywidth\linewidth]{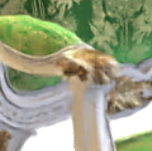} &
  \includegraphics[width=\mywidth\linewidth]{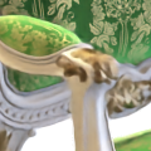} &
  \includegraphics[width=\mywidth\linewidth]{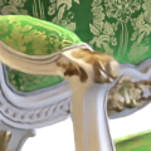} &
  \includegraphics[width=\mywidth\linewidth]{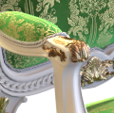} \\
  \begin{small}\makecell[c]{w/o Joint}\end{small} &
  \begin{small}\makecell[c]{K=3}\end{small} &
  \begin{small}Ours\end{small} &
  \begin{small}GT\end{small} 
 \end{tabular}\vspace{-0.2cm}
 \caption{{\bf Ablation study.} Impact of different training configurations has been shown in above visual examples. 
 }
 \label{fig:ablation}
 \vspace{-0.3cm}
\end{figure}

\begin{table}[t]
\small
\centering
\begin{tabular}{@{}lccc@{}}
\toprule
$K$: \#  Feature channels    &    3    &   6   &    9 \\ \midrule
PSNR(dB)                   &    32.53  &  33.14    &  33.21     \\ 
SSIM                    &    0.959  &  0.965    &  0.966     \\ 
LPIPS                   &    0.042  &   0.035    &   0.034   \\ \bottomrule
\end{tabular}
\vspace{-0.2cm}
\caption{\textbf{Comparison of Number of Feature Channels} on Chair.}
\label{fc}
\end{table}

\boldparagraph{Joint Training}
The necessity of joint training is validated in Tab.~\ref{jt}. The method without joint training means the parameters of neural feature fields are frozen when training the neural renderer. Joint training helps neural feature fields generate more expressive features compatible with following neural renderer to synthesize higher-quality textures. 
In Fig.~\ref{fig:ablation}, we show visual examples of different training configurations.
\begin{table}[t]
\small
\centering
\begin{tabular}{@{}lccc@{}}
\toprule
                        & PSNR & SSIM & LPIPS \\ \midrule
Ours w/ joint training  & 33.05    &  0.965     &  0.038     \\ 
Ours w/o joint training & 31.98    &  0.957    &  0.053 \\ \bottomrule
\end{tabular}
\vspace{-0.2cm}
\caption{\textbf{Joint training} on Chair.}
\label{jt}
\end{table}

\boldparagraph{Memory Usage of Neural Feature Fields}
In Tab.~\ref{hashlen}, we compare the visual quality and memory usage when neural feature fields is configured with different length of hash table, \ie, learnable parameters. 
We find that even when the length of the hash table is greatly reduced, the final visual quality is only slightly affected while the memory usage is significantly reduced.
\begin{table}[t]
\small
\centering
\begin{tabular}{@{}lcccc@{}}
\toprule
\makecell[l]{Length of\\Hash   Table} & PSNR(dB) & SSIM & LPIPS & Mem.(MB) \\ \midrule
$2^{19}$  & 33.05     &  0.965
    &  0.038     &  29.2      \\ 
$2^{16}$  & 33.03     &  0.961
    &  0.039     &     8.1   \\ 
$2^{14}$  & 32.70     &  0.953
    &  0.039     &  5.0    \\ \bottomrule
\end{tabular}
\vspace{-0.2cm}
\caption{\textbf{Memory Usage of Neural Feature Fields} on Chair.}
\label{hashlen}
\end{table}

\section{Conclusion}
We propose a new perspective for NeRF rendering acceleration by considering the smooth viewpoint trajectory during interaction. The main idea is to supersample the image rendered at the current viewpoint by taking preceding low-resolution features and depths. The experiments show that our method achieves real-time even when rendering 1080P images.
As a limitation, training the 2D neural renderer is time-consuming. Moreover, we need to train a specialized neural renderer from scratch for each scene individually. In the future, we plan to train a neural renderer that generalizes well on different scenes only after short-time fine-tuning or even without fine-tuning.

{\small
\bibliographystyle{ieee_fullname}
\bibliography{bibliography_long,bibliography,bibliography_custom}
}

\end{document}